\documentclass[letterpaper, 10 pt, conference]{ieeeconf}

\IEEEoverridecommandlockouts

\usepackage{cite}
\usepackage{amsmath}
\usepackage{amssymb}
\usepackage{pifont}
\usepackage{todonotes}
\usepackage{bm}
\usepackage{nicefrac}
\usepackage[noend]{algorithmic}
\usepackage{algorithm}
\usepackage{booktabs}
\usepackage{enumerate}
\usepackage{hyperref}

\usepackage{siunitx}
\usepackage{acro}
\acsetup{patch/maketitle=false}
\DeclareAcronym{LQR}{
    short = LQR,
    long = linear-quadratic regulator
}

\DeclareAcronym{CNN}{
    short = CNN,
    long = convolutional neural network
}
\DeclareAcronym{MLP}{
    short = MLP,
    long = multilayer perceptron
}
\DeclareAcronym{LSTM}{
    short = LSTM,
    long = long short-term memory
}

\DeclareAcronym{RL}{
    short = RL,
    long = reinforcement learning
}
\DeclareAcronym{DRL}{
    short = DRL,
    long = deep reinforcement learning
}
\DeclareAcronym{DroQ}{
    short = DroQ,
    long = dropout Q-function
}

\DeclareAcronym{ECU}{
    short = ECU,
    long = electronic control unit
}
\DeclareAcronym{BEV}{
    short = BEV,
    long = battery electric vehicle
}
\DeclareAcronym{TM}{
    short = TM,
    long = thermal management
}
\DeclareAcronym{TS}{
    short = TS,
    long = thermal system
}
\DeclareAcronym{TSS}{
    short = TSS,
    long = thermal system scenario
}

\DeclareAcronym{MAE}{
    short = MAE,
    long = mean absolute error
}
\DeclareAcronym{RMSE}{
    short = RMSE,
    long = root mean squared error
}
\DeclareAcronym{MTV}{
    short = MTV,
    long = mean total variation
}

\newcommand{\matlab}{\textsc{MATLAB}\textsuperscript{\textregistered}}
\newcommand{\simulink}{\textsc{Simulink}\textsuperscript{\textregistered}}

\newcommand{\rom}[1]{\expandafter{\romannumeral #1\relax}}
\newcommand{\Rom}[1]{\uppercase\expandafter{\romannumeral #1\relax}}

\makeatletter
\newcommand\fs@betterruled{%
  \def\@fs@cfont{\bfseries}\let\@fs@capt\floatc@ruled
  \def\@fs@pre{\vspace*{0.75em}\hrule height.8pt depth0pt \kern2pt}%
  \def\@fs@post{\kern2pt\hrule\relax}%
  \def\@fs@mid{\kern2pt\hrule\kern2pt}%
  \let\@fs@iftopcapt\iftrue}
\floatstyle{betterruled}
\restylefloat{algorithm}
\makeatother

\title{
\vspace{-3.5em}\footnotesize This work has been accepted at the IEEE for publication.\\
Copyright may be transferred without notice, after which this version will no longer be accessible.\\\vspace{5.25em}%
\LARGE \bf
Scenario-based Thermal Management Parametrization \\ Through Deep Reinforcement Learning
}

\author{Thomas Rudolf$^{\,*\,1}$, Philip Muhl$^{\,*\,2}$, Sören Hohmann$^{3}$ and Lutz Eckstein$^{4}$%
\thanks{*These authors contributed equally.}%
\thanks{$^{1}$Thomas Rudolf is with the department for Software Calibration and Diagnosis, Porsche Engineering Group GmbH, 71287 Weissach, Germany,
        {\tt\small thomas.rudolf@porsche-engineering.de}}%
\thanks{$^{2}$Philip Muhl is with the department for Thermal Management Systems, Dr. Ing. h.c. F. Porsche AG, 71287 Weissach, Germany,
{\tt\small philip.muhl2@porsche.de}}%
\thanks{$^{3}$Sören Hohmann is head of the Institute of Control System (IRS), Karlsruhe Institue of Technology (KIT), 76131 Karlsruhe, Germany,
        {\tt\small soeren.hohmann@kit.edu}}%
\thanks{$^{4}$Lutz Eckstein is head of the Institute for Automotive Engineering (ika), RWTH Aachen, 52056 Aachen, Germany,
        {\tt\small lutz.eckstein@ika.rwth-aachen.de}}%
}

\begin{document}

\maketitle

\thispagestyle{empty}
\pagestyle{empty}

\begin{abstract}
The thermal system of battery electric vehicles demands advanced control.
Its thermal management needs to effectively control active components across varying operating conditions.
While robust control function parametrization is required, current methodologies show significant drawbacks.
They consume considerable time, human effort, and extensive real-world testing.
Consequently, there is a need for innovative and intelligent solutions that are capable of autonomously parametrizing embedded controllers.
Addressing this issue, our paper introduces a learning-based tuning approach.
We propose a methodology that benefits from automated scenario generation for increased robustness across vehicle usage scenarios.
Our deep reinforcement learning agent processes the tuning task context and incorporates an image-based interpretation of embedded parameter sets.
We demonstrate its applicability to a valve controller parametrization task and verify it in real-world vehicle testing.
The results highlight the competitive performance to baseline methods.
This novel approach contributes to the shift towards virtual development of thermal management functions, with promising potential of large-scale parameter tuning in the automotive industry.

\end{abstract}

\begin{figure}[!t]
    \centering
    \vspace{0.5em}
    \includegraphics[width=0.99\linewidth]{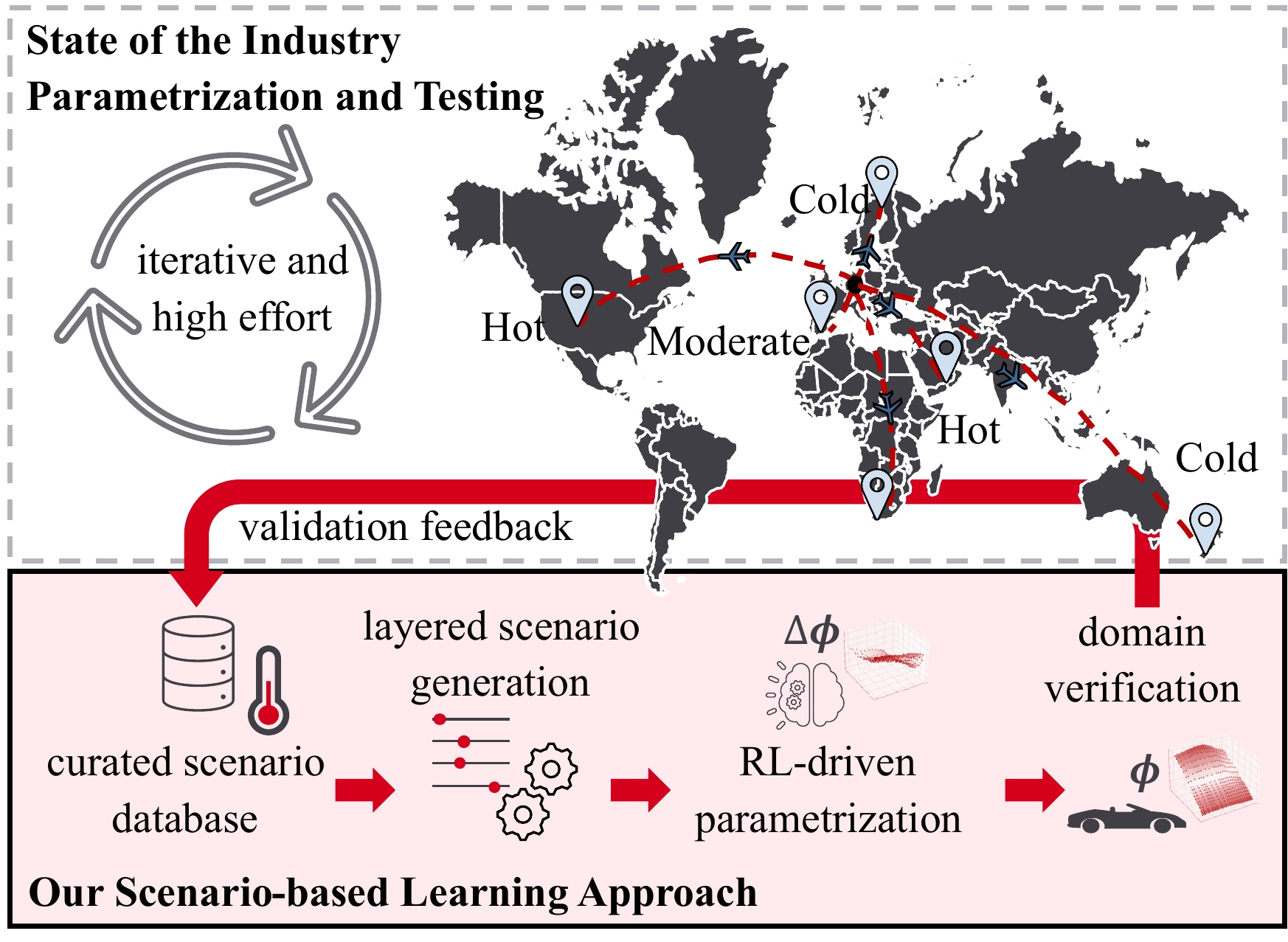}
    \vspace{-1.75em}
    \caption{The current thermal system development process typically requires intensive test iterations in climatic conditions around the world. Our novel approach streamlines the control function parametrization task for reduced testing iterations. We leverage a diverse thermal system scenario generation that drives a learning parametrization agent.}
    \label{fig:visual_abstract}
    \vspace{-1.5em}
\end{figure}
\section{Introduction}\label{sec:introduction}
Modern \acp{BEV} comprise sophisticated \acp{TS}.
The \ac{TS} plays a pivotal role in balancing the \ac{BEV} design challenge with its opposing objectives of energy efficiency, tractive performance, and thermal comfort.
The \ac{TM} addresses the derived operational requirements and oversees the energetically coupled refrigerant and cooling circuits.
Their operation strategy and control pose a large-scale parameter optimization problem in order to provide sufficient thermal power for each component's condition~\cite{Lajunen.2018}.

Tuning the feedback controllers involved in the \ac{TS} demands considerable efforts.
Current virtual parametrization procedures in the automotive industry include gradient-based or global optimization methods that are limited to few parameters and particular test cases~\cite{Knodler.2023}.
To verify vehicle's hardware and software interactions, further improvements and tests are conducted at the real system.
It is validated in iterations and in various climate domains around the world, see Fig.~\ref{fig:visual_abstract}.
However, this is not scalable and falls short in representing the final customer usage patterns.
Thus, virtual development becomes imperative for sustainable verification and validation loops~\cite{Lajunen.2018, Muhl.2023}.

A survey of expert \ac{TS} calibration engineers, see appendix~\ref{app:survey}, highlights two pressing requirements: a) enhanced automation in the parametrization task, and b) adaption to \ac{TM} complication and changing \ac{TS} configurations. 
We identify the need for an intelligent methodology to increase real-world scenario coverage during \ac{TM} optimization.
With the importance around recurring and adaptive creation of initial parameter sets for process efficiency, we investigate learning methods and \ac{TSS} generation.

Previous work outlines a scenario development method for highly automated driving~\cite{Riedmaier.2020}.
The method has been transferred to \ac{TS} verification challenges~\cite{Muhl.2023}.
To minimize manual efforts, the relevance of data-driven approaches within automotive research and development has continuously increased~\cite{Knodler.2023, Puccetti.2021, McClement.2022}.
However, due to cost-effectiveness and traceability in safety-critical applications, explicit embedded logic will likely endure~\cite{Muhl.2022, McClement.2022}.
Therefore, we limit this paper's scope to intelligent parameter optimization methods for embedded functions instead of learning direct control.

Solutions with fixed-structure control systems and its complications have been widely applied.
Gain-scheduling PID control is used for nonlinear problems, and its control parameter dependencies account for exogenous influences on the system dynamics~\cite{McClement.2022, Rudolf.2023}.
Further effort is required whenever subsystems are changed, or vehicle variants are introduced.
Today's methods for control function optimization still struggle to handle the high dimensionality of the parameter space.
The industry lacks sufficient approaches to deal with extensive search spaces while satisfying the flexibility and speed, demanded by the development cycles~\cite{Knodler.2023, Rudolf.2023}.
Our paper focuses on solving these fixed-structure feedback-control parametrization tasks autonomously.

In~\cite{Muhl.2022, Rudolf.2023, Roulet.2024}, autonomously learning agents have been investigated to address aforementioned challenges.
In these works, data-driven parameter tuning processes are formulated as sequential decision-making tasks that enables \ac{DRL} as suitable approach.
Though, the training data generation mainly focused on existing measurements from historic system operation.
Furthermore, only a small parameter subset per decision or operating region could be regarded due architectural inflexibility.

Further industrial applications have recently been explored, such as vehicle velocity tracking control~\cite{Puccetti.2021}, parameter tracking under nonlinear system dynamics~\cite{McClement.2022, Rudolf.2022}, or chip design~\cite{Roy.2021, Mirhoseini.2021}.
In~\cite{Roy.2021} and~\cite{Mirhoseini.2021}, the authors approach matrix-like circuit and chip design using efficient neural networks, often used in image processing.
Similarly, embedded implementations involve matrix-like parameter look-ups.

In this work, we introduce a methodology that integrates the scenario method as technique for \ac{TS} excitation and integrate it into a learning parametrization process for feedback-control.
Our key contributions are as follows:
\begin{enumerate}
    \item we propose a novel image-based \ac{ECU} parameter representation for efficient processing,
    \item we combine the TS scenario generation~\cite{Muhl.2023} with a \ac{DRL} agent for control function parametrization, and
    \item we demonstrate the applicability to a \ac{TM} valve controller and evaluate it in the real-world \ac{BEV}.
\end{enumerate}
We verify the resulting parameter sets on control performance metrics in three scenarios against baselines.
Finally, we conclude with a discussion of our findings and potential future advancements.

\section{Background}\label{sec:background}

\subsection{Thermal System Overview}\label{sec:thermal_system}
\begin{figure}[!t]
    \centering
    \vspace{0.4em}
    \includegraphics[width=0.99\linewidth]{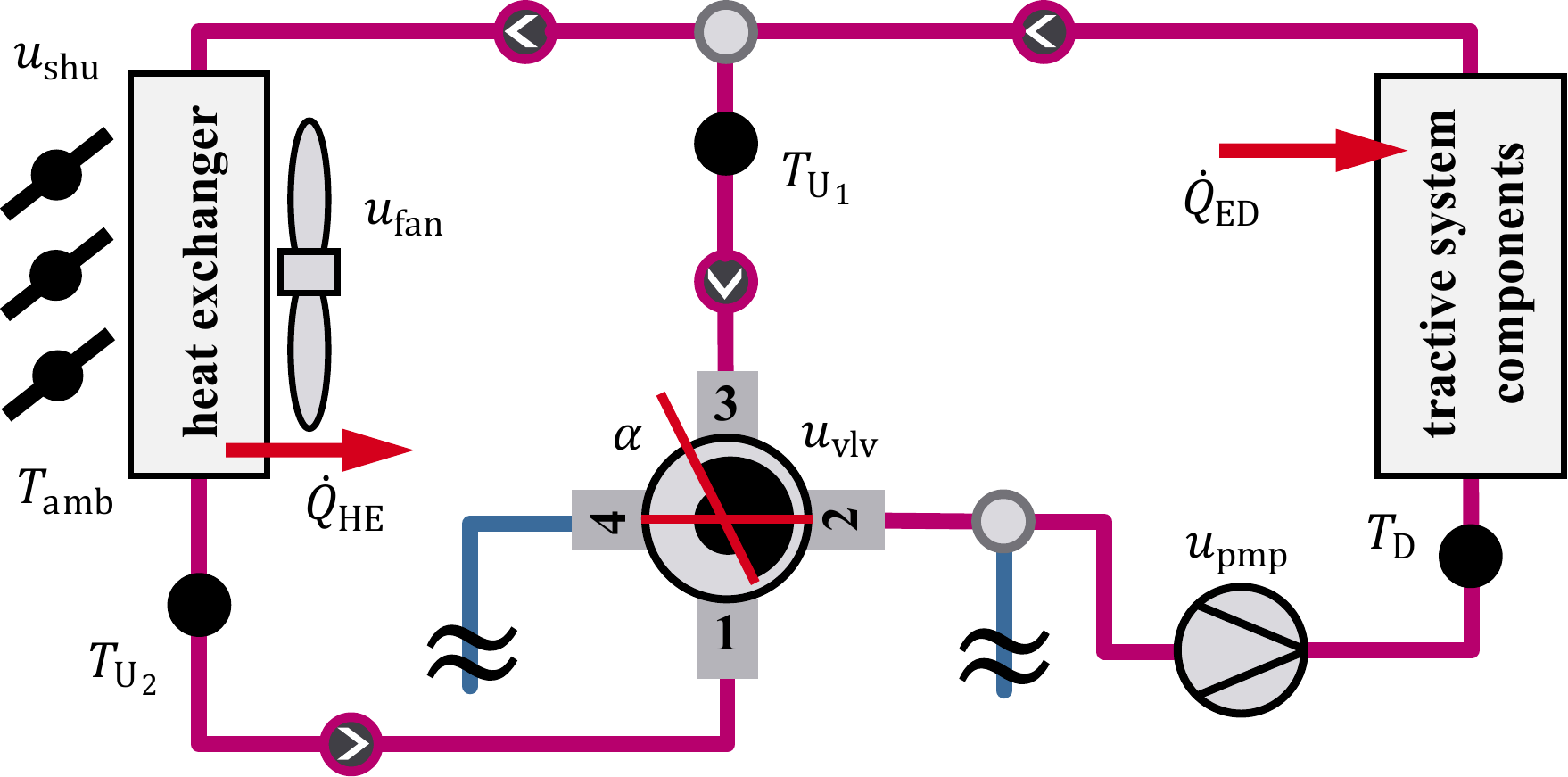}
    \vspace{-1.5em}
    \caption{System schematic contains a continuous rotational four-way mixing valve, a coolant pump, a fan, and a radiator shutter, as the thermal system actuators. The tractive system is the dominant heat source, while the heat exchanger represents the primary heat sink to ambient.}
    \label{fig:system_schematic}
    \vspace{-1.5em}
\end{figure}

Intelligent use of available heat has a positive impact on the primary energy demand of \acp{BEV}.
Advanced \acp{TS} use valves to couple the different sub-circuits in relevant operation scenarios and leverage efficiency gains through optimal use of available heat~\cite{Muhl.2022}.
The topological schematic of the \ac{TS} in this study is depicted in Fig.~\ref{fig:system_schematic}.
The continuous actuation of the mixing valve $u_{\textrm{vlv}}$ is proportional to its rotary piston angle $\alpha$, and controls the downstream temperature $T_{\textrm{D}}$ by mixing two enthalpy streams with the upstream temperatures $T_{\textrm{U}_1}$ and $T_{\textrm{U}_2}$ toward a set-point temperature.
The downstream temperature $T_{\textrm{D}}$ depends on both upstream temperatures, the valve angle $\alpha$, and coolant flow rate in the system.
An electric pump imposes a flow rate $\dot{V}$ on the hydraulic system proportional to the hydraulic plant characteristics $k_{\textrm{hyd}}$ and the pump actuation $u_{\textrm{pmp}}$.
Further, the electric fan speed $u_{\textrm{fan}}$ and radiator shutter angle $u_{\textrm{shu}}$ control the air volume flow $\dot{V}_{\textrm{air}}$ over the heat exchanger proportional to the vehicles speed $v_{\textrm{veh}}$.
The heat flux $\dot{Q}_{\textrm{ED}}$ of electric drive tractive components is proportional to the power $P_{\textrm{ED}}$, vehicle speed $v_{\textrm{veh}}$, and the thermal and electrical operational states $\left\{ \xi_{\textrm{th}}, \xi_{\textrm{el}} \right\}$.
Since the system heat rejection $\dot{Q}_{\textrm{HE}}$ is nonlinear w.r.t. the air volume flow $\dot{V}_{\textrm{air}}$, the partial coolant flow rate $\dot{V}$, the temperature difference $\Delta T_\mathrm{amb}$ between $T_{\textrm{U}_1}$ and the ambient temperature $T_{\textrm{amb}}$.
The nonlinear \ac{TS} dynamics can be described as a function of relevant actuator controls $\left\{u_{\textrm{pmp}}, u_{\textrm{vlv}}, u_{\textrm{fan}}, u_{\textrm{shu}} \right\}$ and the potential variables, result in:
\begin{align}
    \dot{T}_{\textrm{D}}     &=  f(\alpha, \dot{V}, T_{\textrm{U}_1}, T_{\textrm{U}_2}, T_{\textrm{D}}), \\  %
    \begin{split}
    \dot{T}_{\textrm{U}_1}    &=  f(\dot{Q}_{\textrm{ED}}, \dot{V}, T_{\textrm{U}_1}) \\
                    &=  f(P_{\textrm{ED}}, v_{\textrm{veh}}, \xi_{\textrm{th}}, \xi_{\textrm{el}}, k_{\textrm{hyd}}, u_{\textrm{pmp}}, T_{\textrm{U}_1}),
    \end{split} \\
    \begin{split}
    \dot{T}_{\textrm{U}_2}    &= f(\alpha, \dot{Q}_{\textrm{HE}}, \dot{V}, T_{\textrm{U}_2}) \\
                    &= f(u_{\textrm{vlv}}, u_{\textrm{fan}}, u_{\textrm{shu}}, v_{\textrm{veh}}, \Delta{T}, k_{\textrm{hyd}}, u_{\textrm{pmp}}, T_{\textrm{U}_2}).
    \end{split}
\end{align}
Based on the operational conditions, $T_{\textrm{D}}$ is mixed via the ports configuration sets $\left\{ 1, 2, 3 \right\}$ or $\left\{ 1, 3, 4 \right\}$ and the temperature mixing function is either implemented in an upstream or downstream flow control.
Similar dependencies for the level of heat rejection can be defined between the system free cut (blue) in Fig.~\ref{fig:system_schematic} induced by connected subcircuits.

The parametrization of the described temperature controller and \ac{TM} represents a difficult resources and time-intensive process~\cite{Lajunen.2018}. 
State of the industry methods predominantly use physical prototypes for testing in all relevant ambient climatic zones~\cite{Muhl.2023}, see Fig.~\ref{fig:visual_abstract}. 
The described system is representative to various enthalpy controllers in \acp{BEV} with similar configurations and system dynamics, yet simply understandable in the discussed form.

\subsection{Controller Parametrization Problem Formulation}
The \ac{TS} is a nonlinear parameter-varying dynamic system:
\begin{equation}\label{eq:lpv}
	\dot{\bm{x}}(t) =
    	\bm{f} \left(
    	\bm{x}(t), \bm{u}(t), \bm{\theta}(\bm{x}, \bm{u}, \bm{w})
    	\bm \right),
\end{equation}
with system states $\bm{x}$, inputs $\bm{u}$, and unknown variant parameters $\bm{\theta}$ depending on exogenous inputs $\bm{w}$.
The exogenous inputs $\bm{w}$ influence the dynamic states $\bm{x}$ not directly but describe other vehicle's subsystem states.
We regard the nonlinear system dynamics $\bm{f}$ as unknown.

The regarded \ac{TS} features a nonlinear valve geometry and hose routing design that influence the dynamic system behavior. 
In addition to its nonlinearity, dead time from \ac{TS} transport delay or actuators complicates the procedure of finding an optimal control law.
Furthermore, changes to the \ac{TS} configuration through different valve positions couple subcircuit dynamics and create coupling dynamics.
By understanding the significance of these valve positions with respect to the thermal flow management within the system, the criticality becomes evident. Each adjustment of the position has a direct influence on the overall system efficiency and responsiveness.
The continuous positioning of the valve, see Fig.~\ref{fig:system_schematic},  results in a) a lower downstream temperature when opening the 1-2 connection with flow via the radiator while increasing the dead time due to an increased transport length. Vice versa, b) opening the 2-3 connection, thus simultaneously closing 1-2, creates a shortcut via $T_{\mathrm{U_1}}$ and reduces the transportation via the radiator, inhibiting a shorter dead time, and cycles heat from the thermal load.
Simultaneously, the time constants depend on the temperature difference between the two upstream fluid temperatures.
Wax thermostats, the previous industry-standard solution, are not able to fulfil today's dynamic requirements as fixed-gain controllers once the vehicle is at operation temperature and do not allow for efficient heat rejection control.

Fixed-structured PID controllers often apply the gain-scheduling technique to handle nonlinear parameter-varying systems~\cite{McClement.2022} in industrial embedded implementations.
Therefore, we assume the gain-scheduling control law~\cite{Rudolf.2023}:
\begin{equation}\label{eq:gain_scheduling}
    \bm{u}(t) = \bm{g}\left(\bm{x}(t), \bm{y}(t), \bm{w}(t), \bm{\phi}(\bm{x}, \bm{y}, \bm{w}) \right)
\end{equation}
with $\bm{g}$ as function of the system states $\bm{x}$, the outputs $\bm{y}$, and mapping functions $\bm{\phi}$ as adaptive gains based on the operating conditions ($\bm{x}, \bm{y}, \bm{w}$).
We investigate a PI structure:
\begin{align}\label{eq:pi_controller}
    \bm{u}(t) = & \phi_\text{P}(e_\text{T}(t), \Delta T_\mathrm{amb}(t)) \cdot e_\text{T}(t) \,+ \\
                & \quad \phi_\text{I}(e_\text{T}(t), \Delta T_\mathrm{amb}(t)) \cdot \int e_\text{T}(\tau) d\tau, \nonumber
\end{align}
with proportional $\phi_\text{P}$ and integral $\phi_\text{I}$ control gains as a function of the control error $e_T(t)=T_\mathrm{D}(t)-T_\mathrm{D,ref}(t)$ and the difference to ambient temperature $\Delta T_\mathrm{amb}=T_\mathrm{U_1}-T_\mathrm{amb}$ which is proportional to the potential heat dissipation via the heat exchanger.
In embedded industrial controllers, the parameter mappings $\bm{\phi}$ are often implemented as multidimensional lookup tables~\cite{Muhl.2022, Rudolf.2022, Rudolf.2023}.

The challenging parametrization task can be formulated as an optimization problem to find the optimal parameter mapping that minimizes the control objective $J \in \mathbb{R}^+$:
\begin{equation}\label{eq:optimization_problem}
	\bm{\phi}^* = \underset{\bm{\phi}}{\mathrm{argmin}}\,J.
\end{equation}
Though, finding a parameter set that is optimal for all individual usages is infeasible.
Instead, the expectation is to find a compromise over a-priori unknown usage that is robust against fatal performance in edge cases.
Referring to the actual parametrization process, we regard the controller parametrization task as a sequential procedure~\cite{Muhl.2022, Rudolf.2023}.
The parameters are iteratively adapted after each assessment of the current parameter set's closed-loop performance $J$.

\subsection{Scenario-based Virtual Development}
The methodology of advanced scenario-based development, testing and verification was first developed for the complex challenges in assisted and autonomous driving systems (ADAS/AD)  in the PEGASUS research project~\cite{Winner.2019}.
It provides a standardized and regulatory approved virtual development, verification \& validation method to grasp possible driving situations to address the non-feasible billions of necessary testing mileage~\cite{Wachenfeld.2016}. 
The core method projects any driving situation onto six standardized layers.
Each layer represents a distinguished set of information relevant to describe a driving situation to its full extent.
Assuming complete layer information, the recombination of individual layer's data could create any possible driving situation~\cite{Riedmaier.2020}.
This approach is useful for virtual development as well as verification \& validation purposes through standardized scenario description and simulation, e.g., via \textit{OpenDRIVE} and \textit{OpenSCENARIO} formats~\cite{Riedmaier.2020}.
The methodology is recently adapted for \ac{TS} development by enriching the layers with additional domain specific information, as a comprehensive approach to formulate \acp{TSS} for novel tools in the \ac{TM} development and verification~\cite{Muhl.2023}.

\subsection{Reinforcement Learning for Parametrization}
Intelligent control through closed-loop feedback interaction increasingly grows in relevance.
Nonlinear system dynamics, high degree of freedom, or black-box problems are favorable domains for \ac{RL} applications.
Autonomously learning agents have also been applied to white-box nonlinear system identification~\cite{Rudolf.2022} and controller tuning~\cite{McClement.2022, Muhl.2022, Rudolf.2023}.

\begin{figure}[!tb]
    \centering
    \vspace{0.5em}
    \includegraphics[width=0.99\linewidth]{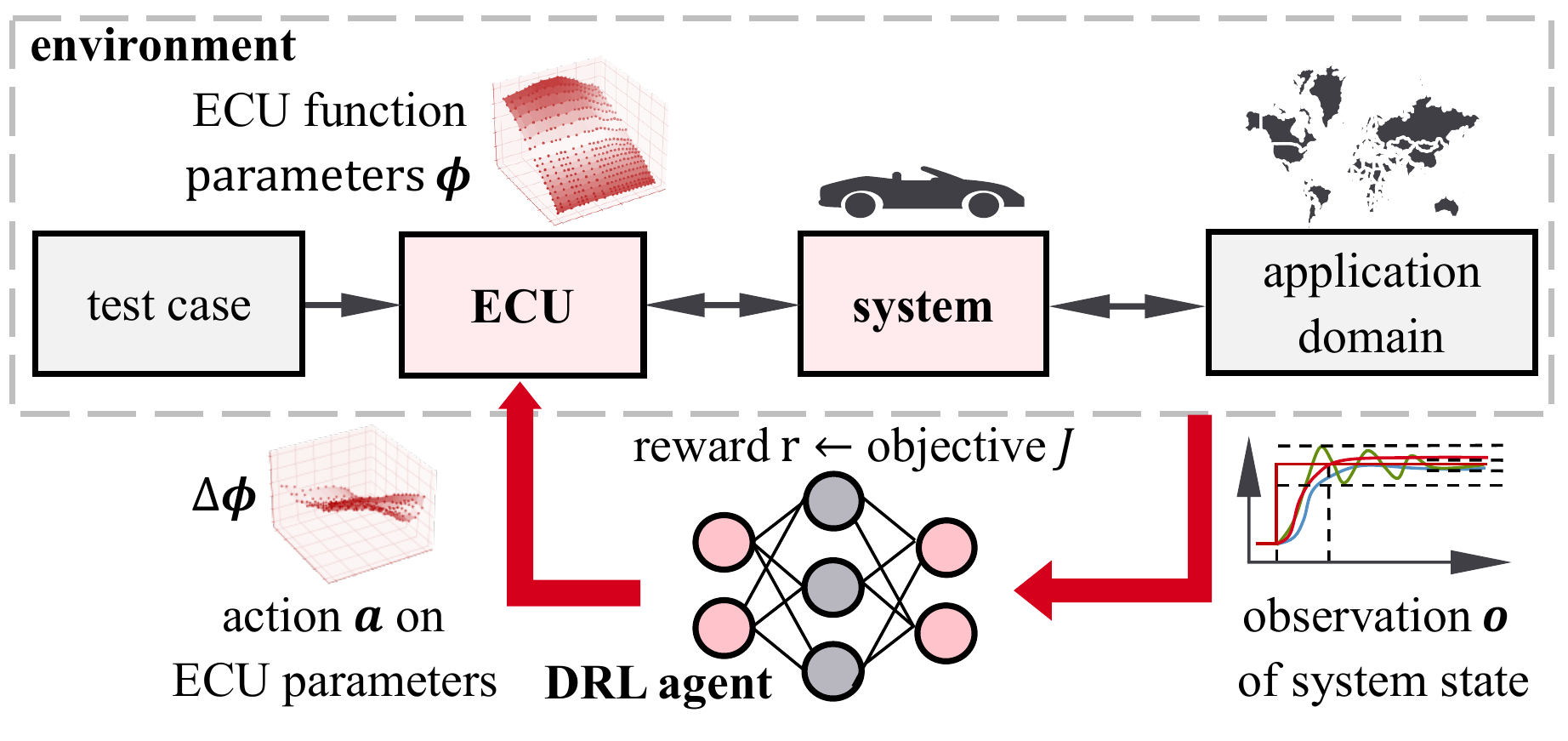}
    \vspace{-2em}
    \caption{Closed-loop controller parametrization diagram, comprising the test-driven ECU parameter evaluation and tuning in a target application domain. A deep reinforcement learning loop interfaces the environment via the observation $\bm{o}$ of the system, adaption of the ECU parameters as actions $\bm{a}$, and rewarding feedback $r$ with respect to the testing objective $J$.}
    \label{fig:rl-loop}
    \vspace{-1.5em}
\end{figure}

The formulated \ac{TM} controller parametrization task can be regarded as a sequential decision-making problem, see~\cite{Sutton.2018, Muhl.2022, Rudolf.2023}. 
As depicted in Fig.~\ref{fig:rl-loop}, static configurations, stationary information and measurable states are observed ($\bm{o}$) from the \ac{TS} in its application domain.
The \ac{RL} environment $\mathcal{E}$ comprises the closed-loop controlled \ac{TS} which is in interaction with the \ac{RL} agent via control performance for different \ac{ECU} parameter sets $\bm{\phi}$.
Subsequently, an action $\bm{a}$ to the control function parameters $\bm{\phi}$ is inferred to tune the \ac{ECU} control function.
Based on the performance $J$, e.g., \ac{LQR} objective, the rewarding feedback $r$ is calculated~\cite{McClement.2022}.
Dense rewards facilitate a steady improvement of the policy $\bm{\pi}$, optimizing the expected sum of discounted rewards with discounting factor $\gamma \in \left[ 0, 1 \right]$~\cite{Sutton.2018}:
\begin{equation}
    \underset{\bm{\pi}}{\text{argmax}}
    E \left\{
        \sum_{n} \gamma^n r_n
    \right\}.
\end{equation}

\section{Our Scenario-Driven RL Approach}\label{sec:approach}
\begin{figure*}[!tb]
    \centering
    \vspace{0.5em}
    \includegraphics[width=0.99\linewidth]{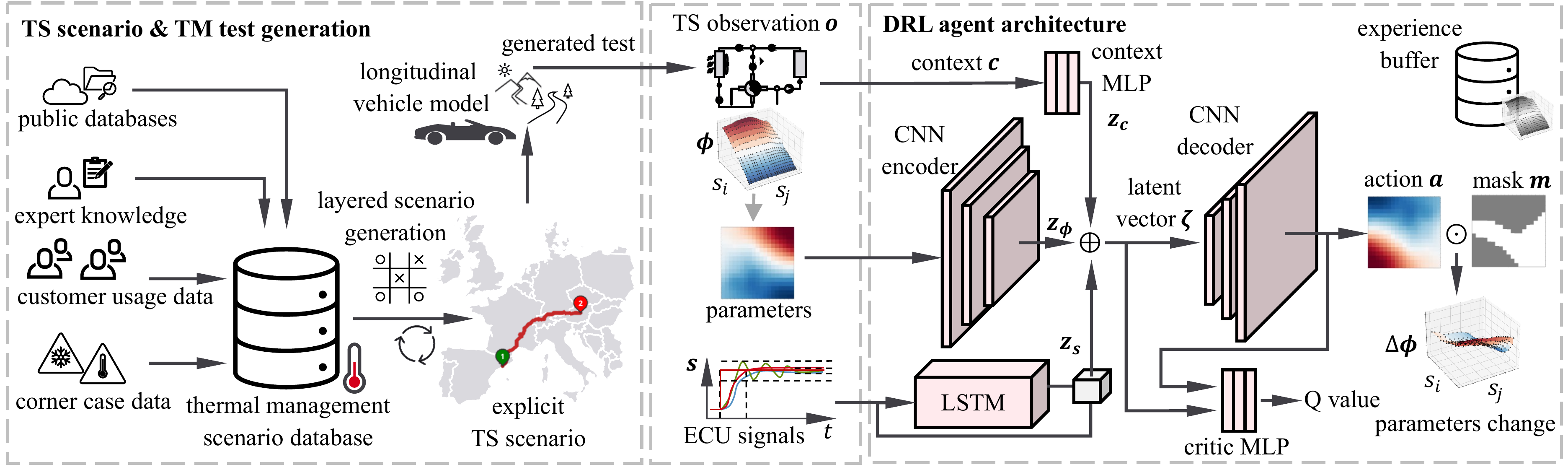}
    \vspace{-1em}
    \caption{Overview of our proposed approach comprising the scenario-based generation of \ac{TSS} (left), the simulation and evaluation of the \ac{TSS} (center), and the architecture of the deep reinforcement learning agent (right).
    The observation $\bm{o}$ comprises the thermal management context information $\bm{c}$, current controller parameters $\bm{\phi}$ as image-like projection, and ECU signals. An encoder-decoder network processes the multi-modal information to predict an action $a$ to the parameters as well as Q values for the learning algorithm. Controller parameter adaptations are masked with respect to operating points.}
    \label{fig:approach_overview}
    \vspace{-1.5em}
\end{figure*}

In order to create versatile training data for the \ac{RL}-based parametrization procedure, we use the adapted scenario methodology introduced by~\cite{Muhl.2023} to create responsive \acp{TSS}, depicted in Fig.\ref{fig:approach_overview} (left).
We collect driving data from real world test drives and customer fleet data as statistical distributions to create a representative vehicle usage scenario database.
Additional expert engineer knowledge and public databases enrich each layer with \ac{TS} information, i.e., charging infrastructure, solar irradiation, and humidity.
We then perform the data allocation to the layer scheme introduced by the adapted scenario methodology.
Through recombination of available information from the scenario database layers in combination with randomized driving routes, we create multiple individual scenarios representing customer-like usage.
A longitudinal dynamics vehicle model calculates the vehicle response trajectory to each scenario, which is used to parameterize and evaluate controller performance.
Furthermore, we explicitly use real world development drive data for the different climatic zones, e.g., measurement drives from Dubai (hot), Sweden (cold), or Spain (moderate), see Fig.~\ref{fig:visual_abstract}. 
The driving data were independently collected over the period of six months with different development vehicles.

\subsection{Methodology for Scenario Generation}
In the following, we present the procedure of the \ac{TS}-adapted scenario methodology to tackle the challenge of understanding and modeling customer vehicle usage for the exemplary valve controller parametrization problem.

The set $\mathcal{M}$ describes vehicle usage data and comprises available customer fleet data and selected development drive data sets. 
We derive layer-focused statistics from $\mathcal{M}$ to describe the usage behavior with respect to \ac{TSS} influencing factors, i.e., mileage per ambient temperature. From this we create a subset $\bm{S}$ with customer-focused metrics, as detailed in~\cite{Muhl.2023}.
A Gaussian-normal distribution fit $\bm{\mathcal{N}} \left( \bm{\mu}_{\bm{S}} , \bm{\sigma}_{\bm{S}} \right)$ for each scenario layer random variable of $\bm{S}$ yields thermal system operation distributions to describe vehicle and \ac{TS} behavior.
To limit the solution space of possible scenarios, we disregard values below the 10\% and above the 90\% behavioral distribution quantiles.
We ensure a robust and safe operation by adding \ac{TS} specialized edge cases, e.g., high loads at extreme temperatures from sub-polar and desert conditions, on top of the broad distributions.
By optimizing the system response for the resulting distributions, development resources are well allocated to improve customer value according to the Pareto-principle.

For the studied valve controller, layers 1, 2, and 5 from~\cite{Muhl.2023} are of particular interest since those hold the primary excitation influences for the control loop.
\textit{Layer~1} information respects the road profile $\left\{ \mathrm{[x,y,z],~road~friction} \right\}$ from map data about the latitude/longitude/altitude position and influences the velocity trajectory.
With respect to vehicular limitations, driver skill and regulatory speed restrictions (see \textit{Layer~2}) the achievable velocity trajectory primarily effects the heat losses from the tractive system.
Further, \textit{layer~5} holds the temperature and humidity ambient conditions, thus affecting heat rejection significantly.

Each draw from the scenario database builds a concrete drive reference with \ac{TS}-related enriched description, denoted as \ac{TSS} for subsequent simulation.
The \acp{TSS} act as system excitations for the overall vehicle with \ac{TS} response based on the parametrizated \ac{TM} in particular.
We use a longitudinal dynamics vehicle model with integrated human driver behavior and thermal system model to provide a simulation result as reference for a virtual test of the \ac{TM} in closed-loop, see Fig.~\ref{fig:approach_overview} (left).
We then simulate batches of \acp{TSS} that result in load, disturbance, and reference \ac{TS} trajectories for different subsystems.
Significant for the simulation of the virtual \ac{TS} is the modeling of thermodynamic and hydraulic phenomena. For that we use existing development system models and the embedded \ac{ECU} software function implementations with the valve controller as the respected tunable part.
In simulated operation, the \ac{TM} changes the cooling circuit configurations for highest efficiency based on the \ac{TS} states $\bm{\xi}$.
This significantly affects the \ac{TS} dynamics and switches between control parameter sets, as described in section~\ref{sec:background}.
The \ac{TS} trajectories are inputs to the \ac{RL} agent and used to calculate the control performance objective.
We limit the scope by disregarding the feedback from the \ac{TS} to the scenario level.
For a given scenario and control parameter set $\bm{\phi}$ we simulate a \ac{TS} dynamic behavior response, see Fig.~\ref{fig:approach_overview} (center), interfacing the subsequent \ac{RL} agent in the parameter adaption loop.

\subsection{Contextual \ac{DRL} for Embedded Controller Tuning}
The following describes our novel contextual \ac{RL}-based controller parametrization agent with its architecture depicted on the right half of Fig.~\ref{fig:approach_overview}.
We describe the \ac{TSS}-driven training algorithm to solve for a contextual parametrization problem based on the obtained \ac{TS} information.

For each decision step about adaptions to the parameters, the agent gathers information about relevant ECU signals and the current parametrization.
The \ac{TS} carries stationary information that describe a context in which the parametrization task needs to be solved, i.e., high-level operational \ac{TM} states $\bm{\xi}$ based on the current \ac{TSS} correlate with significant diverting system dynamics and subsystem control references.
Accordingly, we regard a context vector $\bm{c}=\left[\xi_{\textrm{th}}, \bm{y}_\mathrm{ref}\right]$ with $\xi_{\textrm{th}}$ as thermal operational state describing the desired \ac{TS} sub-circuit configuration and therefore possible valve control piston positions and $\bm{y}_\mathrm{ref}=T_{\mathrm{D,ref}}$ as the downstream temperature reference, desired by the high-level \ac{TM}.
Additionally, we observe measured signals $\bm{s}(n)$ as time-discrete sequences within a windowed interval, with samples $n \in \left[0, N\right]$.
For the valve controller, we regard $\bm{s}(n) = \left[ T_{\mathrm{D}}(n), e_T(n), \Delta T_\mathrm{amb}(n), u_\mathrm{vlv}(n) \right]$.
Last, we need to observe the current tunable parameters $\bm{\phi}$ in~(\ref{eq:pi_controller}).

We reinterpret the embedded implementation of varying control gains in the form of look-up tables as $M_i \times M_j$-sized image-like matrices.
Given the operating states $\left\{ \xi_{\textrm{th}}, \xi_{\textrm{el}} \right\}$, one of two controllers with a PI parameter set is dynamically active and regarded, for a total of \num{100} parameter values.
The matrices $\bm{\phi}$ are dependently evaluated on the current ECU signals, namely $s_i=e_T$ and $s_j=T_\mathrm{amb}$ over their $i, j$-axes with each matrix of size $M=M_i=N_j=5$.
Consequently, we retrieve $C=2$ channels of stacked parameter values, comparable to colour-based channels of an images.
For generalized applicability of the \ac{DRL} architecture, we upsample and interpolate the matrices to $\bm{\tilde{\phi}}$ of common quadratic size $\tilde{M}=8$.
We derive a similar approach for look-ups curves (1D) or cubes (3D) used in the industry.
Finally, we consider the joint \ac{DRL} agent observation $\bm{o} = [ \bm{c}, \bm{\tilde{\phi}}, \bm{s}(n) ]$.

Based on the observation $\bm{o}$, we formulate an \ac{LQR}-inspired objective as a dense reward function:
\begin{equation}\label{eq:reward}
    r_J = - \frac{1}{N} \sum_{n=0}^{N-1} b_1 \left( \sqrt{e_T(n)} + \cdot e_T^2(n) \right) + b_2 \cdot u_\mathrm{vlv}^2(n),
\end{equation}
with design choices $b_1=25$, and $b_2=0.1$ in this work. 
We quadratically penalize large errors and the required cooling power through wide open valve positions but also stimulate the sensitivity at small control errors with the root term.
Unstable controller parameters result in a extreme negative reward that may lead to divergence during training.
Therefore, we clip the negative rewards to a lower limit and clip the positive controller parameters to an upper limit.

The \ac{DRL} architecture in Fig.~\ref{fig:approach_overview} (right) processes the observation $\bm{o}$ and derives the action $\bm{a}=\Delta\bm{\tilde{\Phi}}$ to tune the parameters $\Phi$ in expectation of subsequent rewards.
Neural networks, hence \ac{DRL}, are predominantly applied to sophisticated \ac{RL} problems to approximate a decision strategy~\cite{Sutton.2018, Hiraoka.2022}.
In parallel, a) the task context is inferred to a latent context vector $\bm{z}_{\bm{c}}$ via a \ac{MLP} with nonlinear activations and dropout layers, b) the signals are processed via recurrent \ac{LSTM} layers with the final output as latent signal vector $\bm{z}_{\bm{s}}$~\cite{Muhl.2022, Rudolf.2023}.
Previous work did not regard the state-spatial dependency of neighboring operating points for parameter-varying systems and adaptive controllers in the observation~\cite{McClement.2022}. 
In contrast, we use a \ac{CNN} feature extractor to encode the channeled parameter matrices into $\bm{z}_{\bm{\phi}}$.
Subsequently, we derive a latent vector $\bm{\zeta} = \left[ \bm{z}_{\bm{c}}, \bm{z}_{\bm{s}}, \bm{z}_{\bm{\phi}} \right]$ by concatenation of the individual latent representations.
Based on $\bm{\zeta}$, the \ac{CNN} decoder infers the latent vector. Alternating convolution blocks and upscaling layers generate the target matrix structure $\tilde{M}$.
With $\bm{\zeta}$, the parallel critic \ac{MLP} estimates the state-action value Q for the \ac{RL} training algorithm~\cite{Hiraoka.2022}.

However, applying the full tuning matrix $\Delta\bm{\phi}$ to the parameter tables would create an infeasible sequence of decisions.
Depending on the \ac{TS} state trajectories of an individual scenario, only subsets of the parameters actively contribute to the closed-loop performance, depending on the parameter table axes $s_{i/j}$.
Action masking contributes to a faster convergence during training, enabling real-world applications of otherwise sample inefficient \ac{RL} algorithms~\cite{Huang.2022}.
Therefore, we introduce binary parameter action masks $\bm{m}\in\{0, 1\}^{M_i \times Mj}$.
During dynamic mask construction, a $1$ is registered for at least one time sample in which the \ac{TS} operates in the vicinal table area.
In contrast to the observation, we downsample and interpolate from $\tilde{M}$ to the original parameter matrices' shapes.
We derive the new controller parameter set for step $k$ via the function $\varphi$, applying a Hadamard product with the masks to the downsampled ($\downarrow$) set of parameter changes:
\begin{equation} \label{eq:sampling}
    \bm{\phi}_k = \varphi \left( \bm{\phi}_{k-1}, \bm{a}, \bm{m} \right) = \bm{\phi}_{k-1} + \downarrow_{M_i,M_j}(\Delta\bm{\phi}_{k}) \odot \bm{m}.
\end{equation}

\subsection{Training of the \ac{DRL} Parametrization Agent}
In this work, we deploy the soft actor-critic algorithm derivative \ac{DroQ} for increased sample-efficiency through multiple $Q$-value updates per step on parallel $Q$-networks~\cite{Sutton.2018, Hiraoka.2022}.
The encoder-decoder actor $\bm{\pi}_{\bm{\psi}}$ and critic $\bm{Q}_{\bm{\omega}}$ networks are parametrized by $\bm{\psi}$ and $\bm{\omega}$, and trained through backpropagation of the policy and Q-objective gradients based on the reward $r$~\cite{Hiraoka.2022}.
Algorithm~\ref{alg:rl_training} describes the training routine for our scenario-based \ac{DRL} controller parametrization task.
It requires a \ac{TSS} subset $\bm{S}$ and a \ac{TS} system dataset $\mathcal{D}_{\bm{\theta},\bm{\phi}}$ for initialization and simulation (lines 2,4).
For each episode of length $K$, we reset the \ac{TS} model and a new scenario $\bm{x}_S$ is sampled via the \ac{TSS} method.
We chose conservative stable parameter sets to initialize each episode.
In each episode step, we sample a new scenario $\bm{x}_S$ via the \ac{TSS} method (lines 5,11) and simulate it with the virtual closed-loop controlled \ac{TS} ($EvalScenario$) which derives the observation and mask (lines 6,12).
The \ac{DRL} actor policy derives the parameter adaption (lines 9, 10) and a reward is calculated ($CalcReward$) w.r.t. \eqref{eq:reward}, and the gathered experience stored in the buffer $\mathcal{R}$ (line 14).

After the successful episodic training from off-policy updates of our \ac{DRL} architecture (lines 15, 16), see Fig.~\ref{fig:results_training}, the converged policy $\bm{\pi}_{\bm{\psi}}$ can be independently inferred with in-distribution observations from real-world systems.

\begin{algorithm}[t]
    \caption{Scenario-based \ac{DRL} agent training routine}\label{alg:rl_training}
    \begin{algorithmic}[1]
        \STATE \textbf{input} usage data subset $\bm{S} \in \mathcal{M}$, parameter dataset $\mathcal{D}_{\bm{\theta},\bm{\phi}}$
        \vspace*{-1.25em}
        \STATE \textbf{initialize} \ac{DRL} agent policy $\bm{\pi}_{\bm{\psi}}$, Q-networks $\bm{Q}_{\bm{\omega}}$, experience buffer $\mathcal{R}$, episode length $K$, counter $k=0$
        \FOR{each episode}
            \STATE Sample ($\bm{\theta},\,\bm{\phi}_0$) from $\mathcal{D}_{\bm{\theta},\bm{\phi}}$
            \STATE Sample scenario batch $\bm{X}_{\bm{S}} \sim \bm{\mathcal{N}} \left( \bm{\mu}_{\bm{S}}, \bm{\sigma}_{\bm{S}} \right)$
            \STATE Sample initial scenario $\bm{x}_{0,\bm{S}} \sim \bm{X}_{\bm{S}}$ 
            \STATE $(\bm{o}_0, \bm{m}_0) \gets EvalScenario\left(\bm{x}_{0,\bm{S}}, \bm{\theta}, \bm{\phi}_0 \right)$
            \WHILE{$k \leq K$}
                \STATE $\bm{a}_k \gets \bm{\pi}_{\bm{\psi}}(\bm{o}_k)$
                \STATE $\bm{\phi}_k \gets \varphi \left( \bm{\phi}_{k-1}, \bm{a}, \bm{m} \right)$
                \STATE Sample new scenario $\bm{x}_{k,\bm{S}} \sim \bm{X}_{\bm{S}}$ 
                \STATE $(\bm{o}_{k+1}, \bm{m}_{k+1}) \gets EvalScenario\left(\bm{x}_{k,\bm{S}}, \bm{\theta}, \bm{\phi}_k \right)$
                \STATE $r_{k} \gets CalcReward(\bm{o}_k, \bm{a}_{k}, \bm{o}_{k+1})$
                \STATE Store $(\bm{o}_{k}, \bm{a}_{k}, r_k, \bm{o}_{k+1})$ in $\mathcal{R}$
                \STATE Update Q-networks $\bm{Q}_{\bm{\omega}}$
                \STATE Update policy encoder-decoder network $\bm{\pi}_{\bm{\psi}}$
                \STATE $k \gets k + 1$
                \IF{$k \geq$ $K$} \STATE End episode \ENDIF
            \ENDWHILE
        \ENDFOR
        \STATE \textbf{output} agent policy $\bm{\pi}_{\bm{\psi}}$
    \end{algorithmic}
\end{algorithm}

\section{Real-World Experiment and Results}\label{sec:results}

To verify our methodology, we apply our \ac{DRL} architecture to a \ac{TS} learning environment, implemented in the \textit{Gymnasium} \ac{RL} standard. 
Referring to section~\ref{sec:background}, the simulation includes our previously discussed cooling circuit.
It comprises the PI valve controller and its adjustable parameter tables from section~\ref{sec:approach}.
We use 32 parallel environments which interface the agent with a \ac{TS} simulation model, using C++ code generated by \matlab \ \simulink.

We build our agent on top of the \textit{PyTorch}-based framework \textit{stable-baselines3}~\cite{Raffin.2021}, using the proposed agent from Fig.~\ref{fig:approach_overview}.
We extend a \ac{DroQ} agent to automatically construct its neural network architecture and to compute the parameter action mask during runtime. 
On the masked policy outputs, the soft action distribution parameters are set to $\bm{\mu}_{a}=\bm{0}$ and highly negative $\log{\bm{\sigma}}_{a}$.
The hyperparameters are given in TABLE~\ref{tab:hyperparameters}, applied beyond the baseline's default values~\cite{Raffin.2021}.

\begin{figure}[tb]
    \centering
        \input{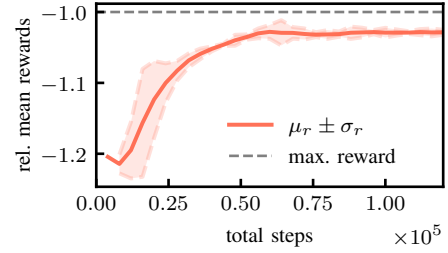}
    \vspace{-1.25em}
    \caption{Agent training curve depicting the rolling mean (window size \num{15} episodes) of the episode rewards and one standard deviation band. The rewards are normalized to the best experienced single scenario performance.}
    \label{fig:results_training}
    \vspace{-1.5em}
\end{figure}

In Fig.~\ref{fig:results_training}, we show the training average reward curve and compare it to the highest reward obtained for a single \ac{TSS}. 
We aim to generate a robust parameter set across the scenario distribution, rather than overfitting a parametrization to a single scenario that produces the maximum reward.
After about \num{60000} training steps at $K=\num{125}$ steps per episode, the episode mean rewards stabilize.
For the real-world verification, we consider the parameter set from the episode with the highest accumulated reward after convergence and transfer it to the ECU.

\begin{figure}[tb]
    \vspace{-0.5em}
    \centering
    \hspace*{-1em}
        \input{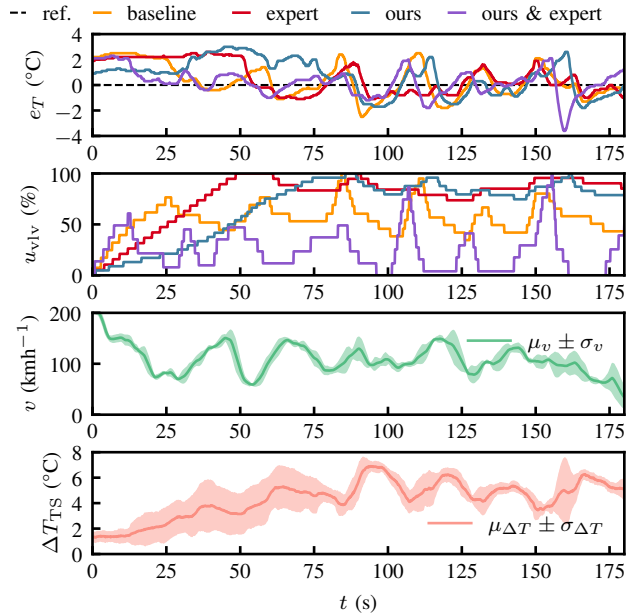}
    \vspace{-2.75em}
    \caption{Recorded ECU signals for scenario~3 over one lap. We consider the parameter sets w.r.t. the control error $e_T$ (top) versus the reference temperature setpoint (black, dashed), the respective valve position $u_\mathrm{vlv}$, and the velocity $v$ and temperature $\Delta T$ profiles.}
    \label{fig:results_signals}
    \vspace{-2em}
\end{figure}

We conduct the verification of our proposed methodology using a real \ac{BEV} sports car. 
Our real-world tests take place at the Nardò Technical Center, Italy.
A location not present in the training dataset but comparable to the climatic conditions represented in a sub-dataset from Spain.
To ensure reproducibility and comparability, each test is executed with the same vehicle and driver.
The initial temperatures are in steady-state and reflect the quasi-constant ambient environment.
We compare four different parameter sets:
\begin{itemize}
    \item baseline: initial parametrization by supplier optimization with functional experience from earlier vehicles,
    \item expert: manual tuning by expert calibration engineer from previous testing under realistic time constraints,
    \item ours: virtual parametrization derived by our approach,
    \item ours \& expert: our parameter set, manually fine-tuned by a calibration engineer on-site.
\end{itemize}
The experiments are real car scenarios:
\begin{enumerate}
    \item oscillating acc-/deceleration between \SI{80}{\kilo\metre\per\hour} \newline and \SI{180}{\kilo\metre\per\hour} into constant velocity of \SI{70}{\kilo\metre\per\hour},
    \item see scenario~1 but into constant velocity \SI{110}{\kilo\metre\per\hour},
    \item sportive Nardò GP track drive, see Fig.~\ref{fig:results_signals}.
\end{enumerate}
The initially oscillating scenarios~1~and~2 evaluate the stabilizing control behavior after intensive thermal excitation into a steady-state heat exchange phase.
The oscillating accelerations up to \SI{180}{\kilo\metre\per\hour} build up a thermal load.
The subsequent constant lower velocity stresses the \ac{TS} and \ac{TM} with significantly reduced heat dissipation via the front heat exchanger.
The delayed fluid transport and decaying heat influx from the \ac{BEV} electric drive challenges the valve controller response.
In contrast, track driving evaluates the dynamic control response to changing \ac{TS} demands.
We prepend an initial lap to condition the thermal system and familiarize the driver with the track conditions, followed by a flying start into the evaluated lap.

We evaluate the recorded \ac{ECU} measurements on the metrics: \ac{MAE} and \ac{RMSE} of the control error $e_T$, the mean square valve actuator energy based on the valve rotation $\dot{u}$, and the \ac{MTV} of the controlled downstream temperature $y=T_\mathrm{D}$ to evaluate the stabilizing behavior.
All metrics are shown in TABLE~\ref{tab:results} with the lowest (best) metric score per test scenario in bold.
When our scenario-based \ac{DRL} agent is involved (ours, ours \& expert), the resulting parameter sets yield superior control error metrics (MAE, RMSE) in all scenarios against the baseline.
Compared to the expert's parameter set, our approach provides better results in the steady-state scenarios.
We consider the wide vehicle usage distributions in our scenario-driven agent as a reason for the slower control response on the track in scenario~3.
However, due to the time advantage via automation with our approach, an a-posteriori fine tuning is possible and sufficient.
In collaboration, ours \& expert achieves the best overall result on the GP track (scenario~3, MAE and RMSE).

\begin{table}[!b]
    \centering
    \vspace{-1em}
    \caption{Evaluation metrics of real-world tests}
    \label{tab:results}
    \begin{tabular}{@{}l l c c c c@{}}
        \toprule
        \textbf{scenario} & \textbf{metric} & \multicolumn{4}{c}{\textbf{parametrization}} \\
        \multicolumn{2}{c}{} & baseline & expert & ours & ours \& expert \\
        \midrule
                & MAE               & 0.849 & 0.290 & \textbf{0.283} & 0.349 \\
                & RMSE              & 1.053 & 0.404 & \textbf{0.384} & 0.455 \\
        scenario~1  & MS$_{\dot{u}}$    & 0.009 & \textbf{0.001} & 0.002 & 0.004 \\
                & MTV$_y$           & 0.021 & \textbf{0.015} & 0.021 & 0.019 \\
        \midrule
                & MAE               & \textbf{0.417} & 0.700 & 0.456 & \textbf{0.417} \\
                & RMSE              & 0.538 & 0.957 & 0.549 & \textbf{0.511} \\
        scenario~2  & MS$_{\dot{u}}$    & \textbf{0.004} & 0.005 &\textbf{ 0.004} & 0.005 \\
                & MTV$_y$           & \textbf{0.017} & 0.021 & 0.020 & \textbf{0.017} \\
        \midrule
                & MAE               & 1.051 & 1.101 & 1.304 & \textbf{0.779} \\
                & RMSE              & 1.305 & 1.359 & 1.496 & \textbf{1.034} \\
        scenario~3  & MS$_{\dot{u}}$    & 0.015 & \textbf{0.007} & 0.010 & 0.015 \\
                & MTV$_y$           & 0.031 & 0.035 & \textbf{0.027} & 0.029 \\
        \bottomrule
    \end{tabular}
\end{table}

Fig.~\ref{fig:results_signals} shows the controller signals during the demanding high-performance scenario~3.
The velocity profile $v$ (green) and resulting coolant temperature rise from the tractive system heat flux $\Delta T_\mathrm{TS} = T_\mathrm{U_1} - T_\mathrm{D}$ are shown in Fig.~\ref{fig:results_signals}.
The temperature difference $\Delta T_\mathrm{TS}$ changes rapidly between maximum power/recuperation phases while showing a phase shift against the velocity profile.
The control error $e_\mathrm{T}$ shows a slow return to the setpoint temperature (dashed) for all parametrizations, except for the collaborative parameter set (ours \& expert, lilac).
This indicates a too slow parameter set for the rapid disturbances to control.
A more aggressive parameter set is necessary to react quickly by opening the valve $u_\mathrm{vlv}$.
However, an overshoot is noticed at \SI{160}{\second}.
Compared to the expert parameter set (red), the ours \& expert combination (lilac) sufficiently reacts to disturbances by dynamically exploiting the full control value range.

In summary, our approach enables a trade-off between performance and robustness across test scenario dynamics.
The results demonstrate the applicability of our approach for real-world parametrization tasks for nonlinear embedded \ac{TM}.
The scenario-based \ac{DRL} approach creates parameter sets that can outperform expert parametrizations.
Nevertheless, the further incorporation of expert knowledge in our agent's parameter sets enhances the control behavior beyond today's levels without our approach.
Real-world testing will remain for validation purposes and presents such opportunity for fine-tuning.
Conclusively, we achieve a virtual and automated process that can reduce development time.

\section{Conclusion}\label{sec:conclusion}
This work presents a novel approach to the parametrization of embedded \ac{TM} valve controllers. 
Leveraging scenario-driven virtual development and image-based parameter representations, our proposed \ac{DRL} agent obtains competitive parameter sets in a fully automated manner.
We demonstrate the validity of learning virtual development processes for the industry applications.
The results show effectiveness to current industry baselines in real-world testing scenarios.
Due to the similarity of enthalpy controllers and system dynamic requirements in \acp{BEV}, the discussed principles can be easily transferred to different controllers within the same vehicle and to different vehicles as well.

Our approach enables streamlined allocation of costly physical prototype resources.
Future work could investigate methodological feedback to the scenario generation within learning curricula.
In addition, the cooperation between autonomous agents and human expert engineers holds great potential for future innovation through distillation of their respective strengths and experience.
Conclusively, the paper highlights the potential for transfers to new application domains and parametrization tasks beyond controller tuning.

\appendix
\subsection{Survey with Expert Calibration Engineers}
\label{app:survey}
\begin{figure}[!htb]
    \vspace{-1em}
    \centering
        \input{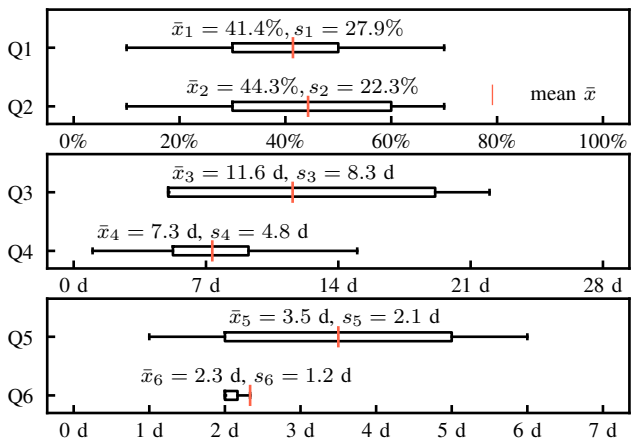}
    \vspace{-2.5em}
    \caption{Box and whisker diagram of the expert survey answers to questions Q$i$ with the respective sample means $\bar{x}_{i}$ and standard deviations $s_{i}$.}
    \label{fig:survey_results}
    \vspace{-1em}
\end{figure}

We conducted a survey with a cohort of expert calibration engineers~($N$=7, 3 junior and 4 senior positions) who regularily parametrize \ac{TM} control systems with $\sim$2000 tunable calibration variables of control architectures as studied in this work. Below is a relevant subset of the survey's questions, and Fig.~\ref{fig:survey_results} shows the respecting statistical evaluations: \newline
Q1: \textit{"What percentage of the total number of tunable parameters can realistically be calibrated?"} \newline
Q2: \textit{"What time saving in the total parametrization process is expected if parameter sets would be generated in seconds?"} \newline
Q3: \textit{"What time effort do you estimate for the parametrization of an exemplary control function for the first \ac{TM} configuration in one climatic zone."} \newline
Q4: \textit{"What additional time effort to Q3 do you estimate for each subsequent \ac{TM} configuration?"} \newline
Q5: \textit{"What time effort do you estimate for the creation of specific test scenario and experiment design?"} \newline
Q6: \textit{"What time effort do you estimate for an initial virtual parametrization in regard to the task in Q3?"}

\subsection{\ac{DRL} agent hyperparameters}
\label{app:hyperparameters}
\begin{table}[!hbt]
    \centering
    \vspace{-1em}
    \caption{}
    \label{tab:hyperparameters}
    \begin{tabular}{@{}lr@{}}
        \toprule
        \textbf{hyperparameter}                 & \textbf{value} \\
        \midrule
        optimizer                               & \texttt{Adam} \\
        learning rate                           & \num{3e-4} \\
        critic updates per steps                & \num{4} \\
        actor updates per steps                 & \num{2} \\
        replay buffer size                      & \num{50e3} \\
        discount factor $\gamma$                & \num{0.98} \\
        activation functions                    & \texttt{ReLU} \\
        upsampled parameter matrix shape        & \num{8}$\times$\num{8} \\
        channels of CNN encoder/decoder (rev.)  & [\num{1}, \num{8}, \num{16}, \num{32}] \\
        kernel size of CNN encoder layers       & [\num{4}, \num{3}, \num{3}] \\
        kernel size of CNN decoder layers       & [\num{2}, \num{2}, \num{2}] \\
        number of LSTM layers                   & \num{2} \\
        number of hidden units per LSTM layer   & \num{256} \\
        number of MLP layers                    & \num{2} \\
        number of hidden units per MLP layer    & \num{256} \\
        \bottomrule
    \end{tabular}
\end{table}

\section*{Acknowledgement}
The authors would like to thank the survey participants in the thermal systems development department at Porsche and the department for software calibration at Porsche Engineering. They further thank the collaborating colleagues at Porsche Engineering Romania as well as Dr. Ing. h.c. F. Porsche AG for providing the test vehicle and infrastructure.

\bibliographystyle{templates/bibliography/IEEEtran}
\bibliography{templates/bibliography/IEEEabrv, indices/references.bib}

\end{document}